\documentclass[10pt,conference]{IEEEtran}
\IEEEoverridecommandlockouts
\usepackage{cite}
\usepackage{braket}
\usepackage{hyperref}
\usepackage{amsmath,amssymb,amsfonts}
\usepackage{algorithmic}
\usepackage{graphicx}
\usepackage{textcomp}
\usepackage{xcolor}
\usepackage{subcaption}
\usepackage{float}
\usepackage{booktabs}    
\usepackage{array}    
\usepackage{tikz}
\usepackage{booktabs}
\usepackage{ragged2e}
\usepackage{caption}
\DeclareCaptionType{listing}[\texttt{Code Block}][List of Listings]
\usetikzlibrary{positioning, shapes.geometric, arrows.meta, fit, backgrounds, calc}

\usepackage[most]{tcolorbox}
\tcbuselibrary{listings}
\usepackage{listings}
\usepackage{xcolor}

\definecolor{merlinRed}{HTML}{F07362}
\definecolor{cellBG}{HTML}{F5F5F5}
\definecolor{merlinBlue}{HTML}{435BEC}

\lstdefinestyle{merlinListing}{
  language=Python,
  basicstyle=\ttfamily\scriptsize,
  keywordstyle=\bfseries,
  commentstyle=\itshape\color{gray!70},
  stringstyle=\color{merlinBlue},
  showstringspaces=false,
  breaklines=true,
  breakatwhitespace=true,
  tabsize=2,
  columns=fullflexible,
  keepspaces=true
}

\newtcblisting{codecellT}[1][]{
  enhanced,
  colback=cellBG,
  colframe=merlinRed,
  boxrule=0.6pt,
  arc=1.5mm,
  left=1.5mm,right=1.5mm,
  top=-0.5mm,bottom=-0.5mm,
  boxsep=0.6mm,
  fonttitle=\bfseries\scriptsize,
  coltitle=black,
  title={#1},
  listing engine=listings,
  listing options={style=merlinListing},
  listing only,
  before skip=0pt,
  after skip=0pt
}

\begin{document}

\title{MerLin: A Discovery Engine for Photonic and Hybrid Quantum Machine Learning
\thanks{This research was supported by MITACS Accelerate (projects IT45761 and IT36314), the UFOQO Project, financed by the French State as part of France 2030, the European Union’s Horizon Europe research and innovation programme under grant agreement No 101130384 (QUONDENSATE) and the QuantERA programme through the project ResourceQ.}
}

\author{
\IEEEauthorblockN{
Cassandre Notton\IEEEauthorrefmark{1}\IEEEauthorrefmark{6},
Benjamin Stott\IEEEauthorrefmark{2}\IEEEauthorrefmark{6},
Philippe Schoeb\IEEEauthorrefmark{1}\IEEEauthorrefmark{3},
Anthony Walsh\IEEEauthorrefmark{2}, 
Grégoire Leboucher\IEEEauthorrefmark{2}\IEEEauthorrefmark{4},\\
Vincent Espitalier\IEEEauthorrefmark{2},
Vassilis Apostolou\IEEEauthorrefmark{2},
Louis-Félix Vigneux\IEEEauthorrefmark{1}\IEEEauthorrefmark{5},
Alexia Salavrakos\IEEEauthorrefmark{2},
Jean Senellart\IEEEauthorrefmark{2}
}

\IEEEauthorblockA{
\IEEEauthorrefmark{1}Quandela Quantique Inc., Montréal, Canada\\
\IEEEauthorrefmark{2}Quandela, Massy, France\\
\IEEEauthorrefmark{3}DIRO \& Mila, Université de Montréal, Montréal, Canada\\
\IEEEauthorrefmark{4}ENS Paris-Saclay, Gif-sur-Yvette, France \\
\IEEEauthorrefmark{5}Département d'informatique, Université de Sherbrooke, Sherbrooke, Canada\\
\IEEEauthorrefmark{6}These authors contributed equally to this work.
}
}

\maketitle

\begin{abstract}
Identifying where quantum models may offer practical benefits in near-term quantum machine learning (QML) requires moving beyond isolated algorithmic proposals toward systematic and empirical exploration across models, datasets, and hardware constraints. We introduce MerLin, an open-source framework designed as a discovery engine for photonic and hybrid quantum machine learning. MerLin integrates optimized strong simulation of linear-optical circuits into standard PyTorch and scikit-learn workflows, enabling end-to-end differentiable training of quantum layers.
MerLin is designed around systematic benchmarking and reproducibility. As an initial contribution, we reproduce eighteen state-of-the-art photonic and hybrid QML works spanning kernel methods, reservoir computing, convolutional and recurrent architectures, generative models, and modern training paradigms. These reproductions are released as reusable, modular experiments that can be directly extended and adapted, establishing a shared experimental baseline consistent with empirical benchmarking methodologies widely adopted in modern artificial intelligence.
By embedding photonic quantum models within established machine learning ecosystems, MerLin allows practitioners to leverage existing tooling for ablation studies, cross-modality comparisons, and hybrid classical–quantum workflows. The framework already implements hardware-aware features, allowing tests on available quantum hardware while enabling exploration beyond its current capabilities, positioning MerLin as a forward-looking co-design tool linking algorithms, benchmarks, and hardware.
\end{abstract}


\section{Introduction and Motivation}
The convergence of quantum computing and machine learning is driving significant advances in both theoretical research and practical applications, with QML offering the potential to accelerate and extend the capabilities of classical algorithms. Among the various quantum platforms, photonic quantum computing is particularly promising due to its scalability, robustness, compatibility with optical communication technologies, and energy efficiency \cite{he_deepquantum_2025,Heurtel2023percevalsoftware,maring_versatile_2024, soret_quantum_2026}. Building on these advantages, photonic QML exploits the bosonic nature of light and high-dimensional multi-mode interference to implement and train machine learning models directly on this unconventional photonic quantum computation model, enabling intrinsic parallelism and efficient exploration of large Hilbert spaces.
Realizing this potential requires frameworks that connect QML models to photonic and other quantum hardware, enabling consistent evaluation across platforms. By embedding trainable quantum circuits into standard AI pipelines, these frameworks support hybrid quantum–classical workflows suited to the near-term regime, where quantum structure can be explored as a source of representations and inductive biases. Consequently, progress in photonic QML depends on scalable, benchmark-driven experimentation rather than isolated algorithmic proposals.

Within this landscape, quantum software frameworks differ markedly in both their level of abstraction and the computational paradigms they support. Hardware-oriented frameworks, such as \texttt{Qiskit} \cite{javadi2024quantum} or \texttt{Cirq} \cite{cirq_developers_2021}, provide mature ecosystems for superconducting processors. By contrast, \texttt{Pulser} \cite{silverio2022pulser} targets neutral-atom platforms, while \texttt{Perceval} \cite{Heurtel2023percevalsoftware} offers state-of-the-art discrete-variable (DV) simulation backends and direct hardware access to Quandela's photonic processors. As for \texttt{Strawberry Fields} \cite{killoran_strawberry_2019} and \texttt{Piquasso} \cite{kolarovszki_piquasso_2025}, they focus on continuous-variable (CV) paradigms, with \texttt{Piquasso} additionally supporting discrete-variable (DV) simulations.

Other software frameworks focus on tools for QML and optimization such as differentiable quantum programming and PyTorch\footnote{PyTorch, the PyTorch logo and any related marks are trademarks of The Linux Foundation.
} integration, like \texttt{PennyLane} \cite{bergholm2018pennylane}, \texttt{TorchQuantum} \cite{wang_quantumnas_2022} and \texttt{Qiskit-Torch-Module} \cite{meyer2024qiskit}. More recently, \texttt{DeepQuantum} \cite{he_deepquantum_2025} emerged as a unified platform bridging qubit circuits, photonic qumodes (Fock, Gaussian, Bosonic backends), and measurement-based quantum computing within PyTorch, reporting GPU-accelerated gradient computation an order of magnitude faster than \texttt{PennyLane} at scale.

This abundance of platforms results in a rich, but fragmented software landscape. Each framework specialises in a particular layer or paradigm, creating silos where algorithms are not portable without significant conversion effort. Most of the QML literature is built on the gate-based paradigm. However, a growing body of work considers algorithms tailored to hardware, within which photonic QML is gaining traction.  A systematic survey found that photonic contributions currently account for $\approx$6\% of QML publications, with most being simulator-based~\cite{notton_establishing_2025}. We confirmed this through independent searches\footnote{arXiv abstract search: 188 photonic QML papers vs.\ 3,592 QML papers; lens.org yields 6.1\%.}.

Recent works demonstrate the potential for photonic QML using a variety of platforms. Gan et al. \cite{gan_fock_2022} explore variational algorithms which project data and training parameters to the high-dimensional Fock space using \texttt{NLopt}~\cite{gan_fock_2022}, while another work ~\cite{yin_experimental_2025} utilizes \texttt{DisCoPy} \cite{de2020discopy} to explore Fock-space encodings for kernel methods. Moreover, additional algorithms, such as quantum reservoir computing implemented with \texttt{Perceval} and \texttt{Keras} ~\cite{sakurai_quantum_2025,rambach_photonic_2025}, and quantum convolutional neural networks using \texttt{QOptCraft} \cite{monbroussou2025photonicquantumconvolutionalneural,gomez2022qoptcraft} have been studied on photonic platforms. Across this surveyed literature, each of these contributions relies on a different ad-hoc software stack, because no existing framework currently combines efficient simulation, integration with ML workflows, noise models and hardware access.

To address this gap, we present \texttt{MerLin}: a framework for photonic QML that builds on \texttt{Perceval}'s optimised DV simulation and hardware access while adding native PyTorch and scikit-learn integration for end-to-end differentiable training. \texttt{MerLin} adopts a photonic-native model, operating directly in Fock space with parametrized linear-optical circuits whose trainable parameters correspond to physical components (e.g., phase shifters and beam splitters described in Section \ref{sec:SLOS}). It exposes measurement primitives aligned with photonic hardware and enables gradient-based training through differentiable strong simulation.

Simultaneously, we designed \texttt{MerLin} as a benchmarking- and reproduction-first platform, providing a unified, hardware-aware platform for implementing, training, and evaluating photonic QML models. Indeed, there has been a need for rigorous benchmarking in the QML community -- called for in \cite{bowles_better_2024,notton_establishing_2025} -- which can be linked to several factors: an extreme heterogeneity in data preprocessing and task formulation in the literature, low code availability preventing independent verification of most results\footnote{Code availability hovers around 27\% for gate-based and 43\% for photonic QML papers.}, and a preference for single-run metrics rather than multi-dimensional evaluations.

We addressed this need by developing a dedicated reproduction framework, designed to enable systematic and ready-to-use replication of published QML works, including reported claims, performance metrics, and experimental analyses, within a controlled software environment built on top of \texttt{MerLin}. As an initial demonstration of this methodology, we reproduced key results from eighteen state-of-the-art photonic and hybrid QML works, spanning expressivity analyses, quantum kernels, reservoir computing, and convolutional architectures. These reproductions validate \texttt{MerLin}’s correctness, and provide a starting point for future contributions.

 Crucially, the objective of this benchmarking effort is not only to identify positive performance gains, but also to understand their origin, and to disentangle improvements due to data preprocessing, model engineering, or optimization strategies from those arising from genuinely new representational or computational mechanisms.

The \texttt{MerLin} framework and the companion reproduction repository are publicly available at \url{https://github.com/merlinquantum/merlin} and \url{https://github.com/merlinquantum/reproduced_papers}.

\section{\texttt{MerLin} architecture}
\begin{figure*}
    \centering
    \includegraphics[width=0.85\linewidth, trim=0.5cm 4.8cm 0cm 1.5cm, clip]{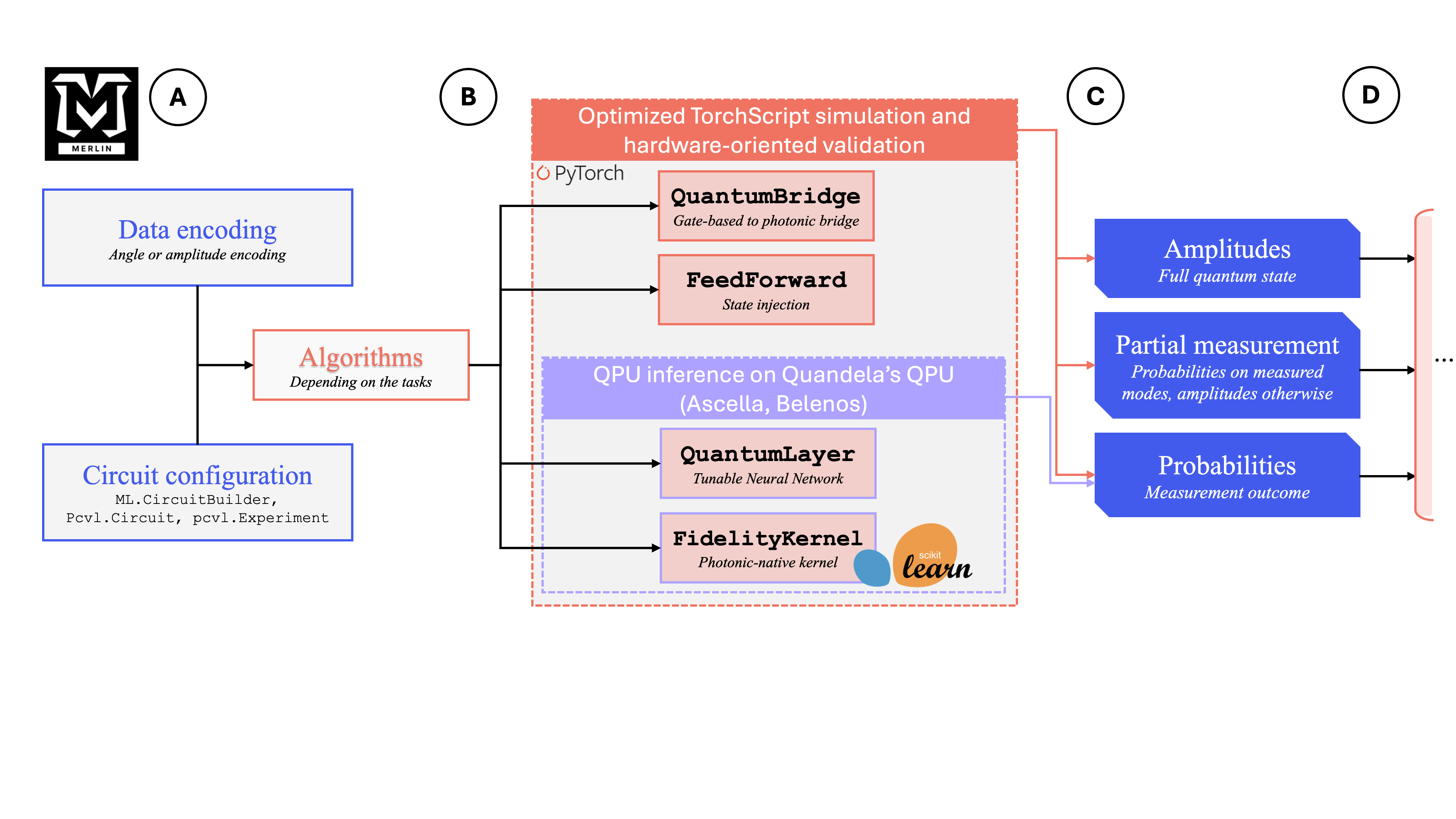}
    \caption{\texttt{MerLin} architecture for photonic QML. \textbf{(A)} Classical data encoding and photonic circuit configuration define the quantum model. \textbf{(B)} \texttt{MerLin} integrates PyTorch-based optimization with photonic-native execution through a logical-to-photonic bridge, differentiable quantum layers, and hardware-oriented simulation, with optional inference on Quandela photonic QPUs. \textbf{(C)} Measurement strategies and detectors behaviour expose full quantum states or partially measured observables. \textbf{(D)} The resulting amplitudes or probability distributions are returned as classical outputs for downstream machine-learning tasks.}
    \label{fig:placeholder}
\end{figure*}
\subsection{Differentiable simulation of Photonic Quantum circuits}
\label{sec:lo_background}
\subsubsection{Essentials of linear optics}
In quantum linear optics, we consider an $n$-photon input state that evolves through an $m$-mode interferometer made of beam-splitters and phase shifters generating superposition and entanglement across photonic modes. This interferometer, or circuit, thus implements an operation that is described mathematically by a unitary matrix $U$ of size $m \times m$. The state at the output of the interferometer is then measured by photon detectors, which register the number of photons in each mode. We denote the input state as $\ket{\mathbf{s}} = \ket{s_1, \dots, s_m}$ where each $s_i$ describes the photon occupancy of the $i$-th optical mode, so that $s_i \in \{0, \dots n\}$ and $\sum_i s_i = n$. The state $\ket{\mathbf{s}}$ is called a Fock state. In lossless regime, the total number of photons $n$ is preserved through the evolution in the interferometer, so the photon detectors measure output arrangements $\mathbf{s}' = (s'_1, \dots s'_m)$, with $\sum_i s'_i = n$. Given a choice of $n$ and $m$, there are $N_{\text{Fock}} = \binom{m+n-1}{n}$ different output states -- the dimension of the Fock space -- and the probability of measuring each of them is determined by the matrix permanent of submatrices of $U$, up to normalization factors \cite{mezher_solving_2023}.

\subsubsection{Strong Linear Optical Simulation}
\label{sec:SLOS}
The Strong Linear Optical Simulation (SLOS) framework \cite{Heurtel_2023}, on which \texttt{MerLin} is built, is a tool that performs \emph{strong simulation}: it computes the exact quantum state after evolution through the interferometer, from which all output probabilities can be extracted, in a time complexity of $\mathcal{O}\left(n\binom{m+n-1}{n}\right)$. While the overall complexity remains proportional to the size of the Fock space, SLOS avoids computing a separate matrix permanent for each output configuration by reusing intermediate results. This provides a significant speedup over previously known simulation methods for many practical cases. The trade-off is memory: SLOS requires $\mathcal{O}\!\left(n\binom{m+n-1}{n}\right)$ 
space to store the complete state vector, which limits practical use to 
approximately $n \lesssim 20$ photons on standard hardware \cite{Heurtel2023percevalsoftware}.

\subsubsection{Circuit optimization with PyTorch}
\label{sec:pytorch-integration}
\texttt{MerLin} integrates SLOS into PyTorch \cite{paszke2019pytorchimperativestylehighperformance} enabling gradient-based optimization of parametrized linear optical circuits via automatic 
differentiation. On photonic hardware, the trainable parameters $\mathbf{\theta}$ correspond to the phases and reflectivities of the optical components (phase shifters and beam-splitters), which together define a parametrized unitary describing the circuit $U(\mathbf{\theta})$.

To accelerate simulation, \texttt{MerLin} provides TorchScript-compiled SLOS primitives built on a precomputed sparse computation graph. SLOS computes the output state iteratively and builds layer-by-layer transitions between intermediate Fock state configurations within a circuit. Starting from a $1$-photon intermediate state, it successively constructs $k$-photon states for $k = 2, \ldots, n$ by applying transition 
rules that depend only on the circuit topology. Specifically, it determines which Fock state configurations at layer~$k-1$ can contribute to which configurations at layer~$k$. This transition structure is determined entirely by the input state, and is independent of the actual unitary matrix entries. 
\texttt{MerLin} exploits this by prebuilding the sparse transition graph once for a given input state, encoding the valid $k \to k+1$ photon transitions as sparse index mappings. During training, only the unitary-dependent coefficients are recomputed at each forward pass, while the graph structure is reused. 

\subsection{The Quantum Layer abstract}
\label{sec:quantum-layer}
\begin{figure*}[t]
\refstepcounter{figure}
\label{fig:vqc_layout}

\centering

\begin{minipage}[t]{0.40\textwidth}
  \vspace{0pt}
  \begin{codecellT}[QuantumLayer initialization and training]
import merlin as ML

builder = ML.CircuitBuilder(n_modes=3)
builder.add_entangling_layer(
    trainable=True, name="W1")
builder.add_angle_encoding(modes=[0, 1], name="x_")
builder.add_entangling_layer(
    trainable=True, name="W2")

layer = ML.QuantumLayer(
    builder=builder,
    input_state=[1, 1, 1],
    measurement_strategy = ML.MeasurementStrategy.probs(
        computation_space = 
            ML.ComputationSpace.FOCK))

optimizer = torch.optim.Adam(layer.parameters(), lr=0.01)
loss_fn = torch.nn.CrossEntropyLoss()

for epoch in range(n_epochs):
    optimizer.zero_grad()
    output = layer(X_train)
    loss = loss_fn(output, y_train)
    loss.backward()
    optimizer.step()
  \end{codecellT}
  \captionof{listing}{QuantumLayer initialization and training using \texttt{MerLin} and a \texttt{PyTorch}-native optimization loop}
  \label{code:merlin_layer}
\end{minipage}
\hfill
\begin{minipage}[t]{0.58\textwidth}
  \vspace{0pt}
  \centering

  \begin{subfigure}[t]{\textwidth}
    \centering
    \includegraphics[width=0.88\textwidth]{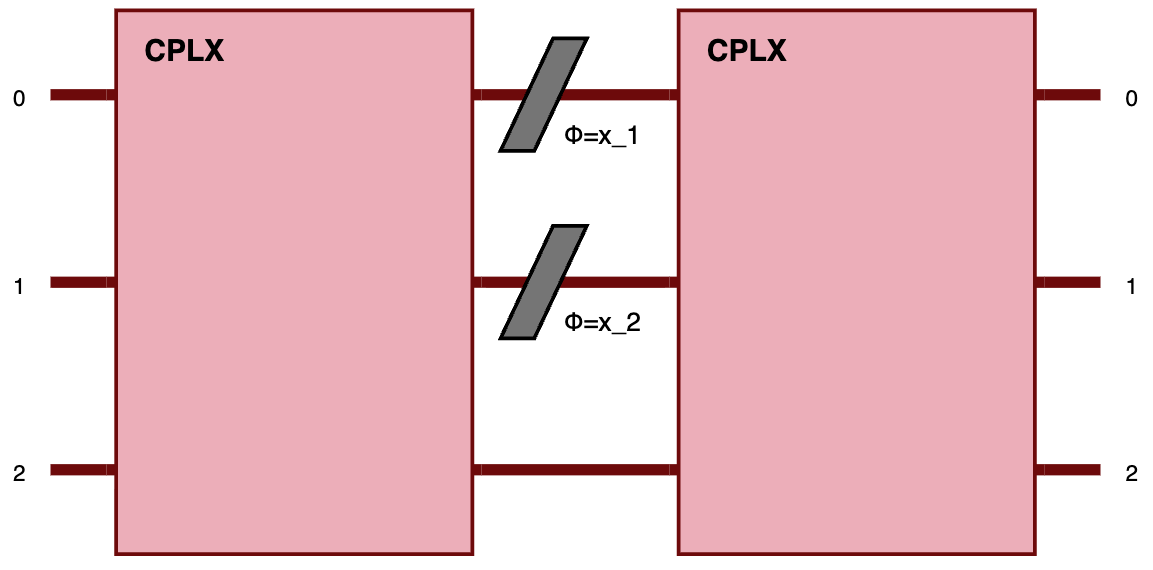}
    \caption{Parameterized linear photonic circuit with $m=3$ spatial modes and single-phase data encoding. The expectation value under PNR or threshold detection decomposes as $\sum_{\omega} c_{\omega} e^{i \omega x}$, where $\omega$ depends on the photon number and $c_{\omega}$ on the trainable circuit and observable.
}
    \label{fig:ql_circuit}
  \end{subfigure}

  \vspace{0.6em}

  \begin{subfigure}[b]{0.64\textwidth}
    \centering
    \includegraphics[width=\textwidth]{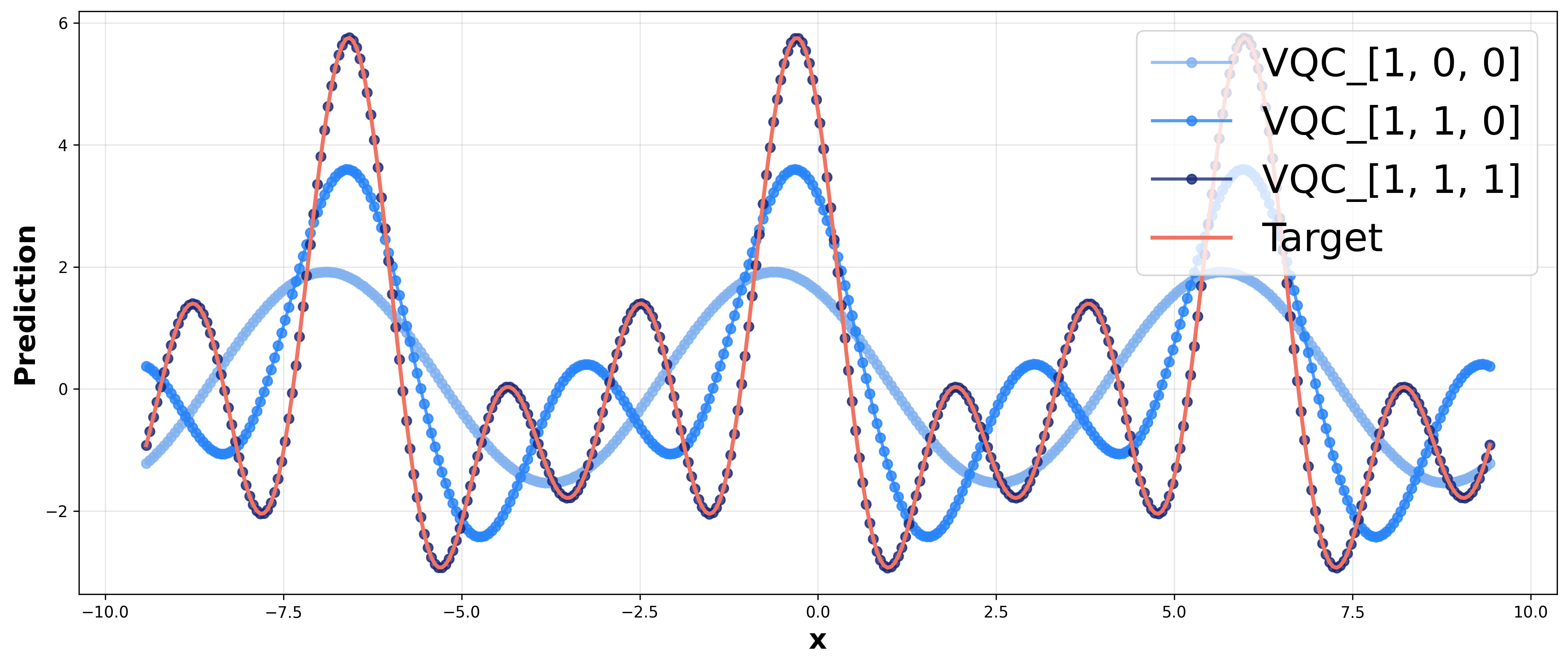}
    \caption{Optimal fits of a degree-three Fourier series $g(x) = \sum_{n=-3}^{3} c_n e^{- i n x}$ obtained with a three-mode photonic circuit using PNR detectors for different input Fock states.
}
    \label{fig:ql_fourier}
  \end{subfigure}
  \hfill
  \begin{subfigure}[b]{0.34\textwidth}
    \centering
    \includegraphics[width=\textwidth]{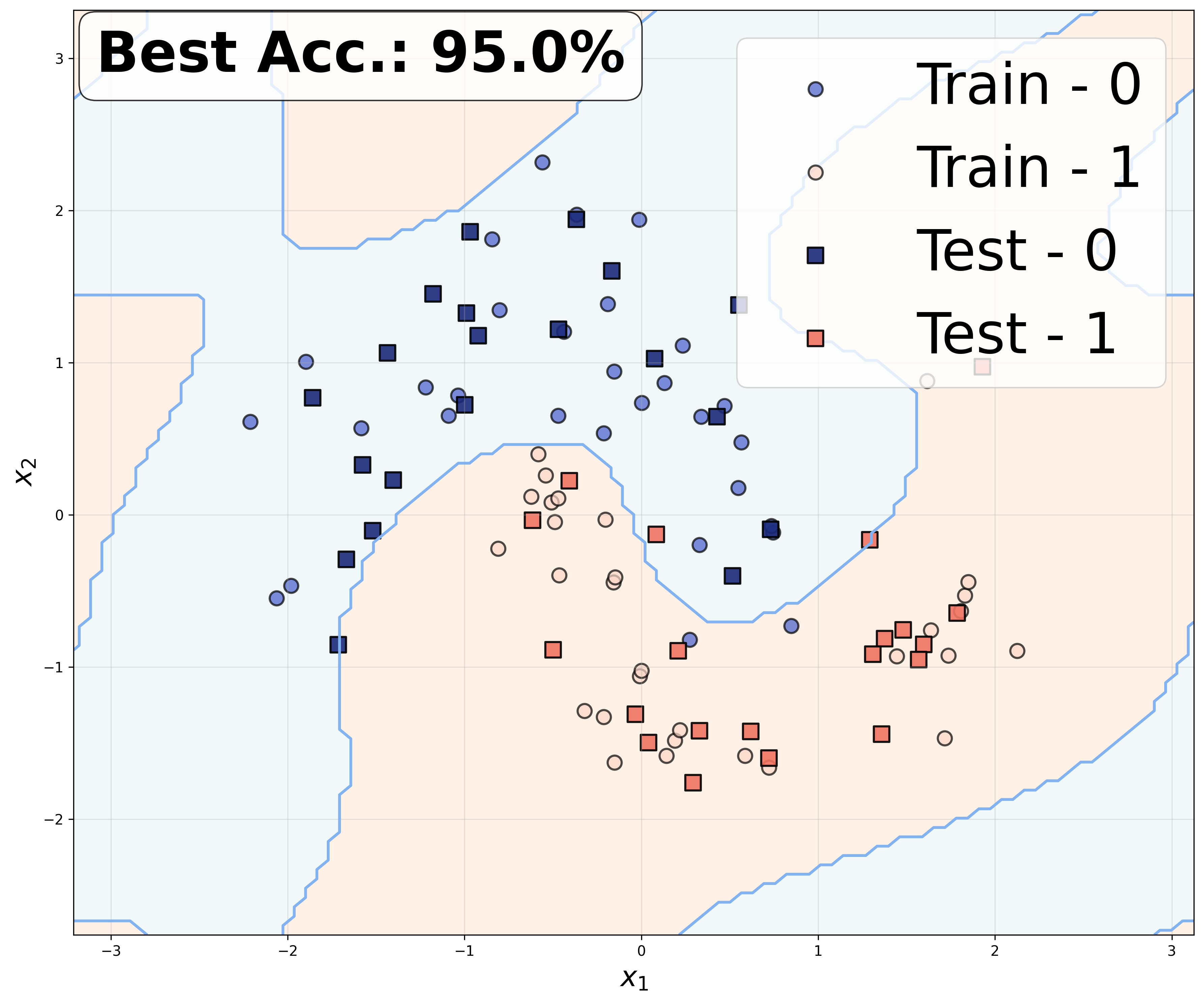}
    \caption{Binary classification on the Moon dataset using the three-mode VQC (\ref{fig:ql_circuit}) with input Fock state \(\lvert 111\rangle\).}
    \label{fig:ql_moon}
  \end{subfigure}

\end{minipage}

\caption*{Figure \thefigure: Overview of a photonic VQC implemented with \texttt{MerLin} and seamlessly integrated into a standard machine learning workflow. \texttt{Code Block} \ref{code:merlin_layer}: PyTorch-native code illustrating the ease of defining, training, and optimizing a \texttt{QuantumLayer}. 
Figure \ref{fig:ql_circuit}: Three-mode linear photonic circuit where classical data are encoded as phase shifts between two trainable universal interferometers. Figure \ref{fig:ql_fourier}: The resulting VQC naturally realizes a truncated Fourier series. Figure \ref{fig:ql_moon}: Despite its simplicity, the same three-mode VQC introduces sufficient nonlinearity to perform binary classification on the moon dataset.}
\end{figure*}

\texttt{MerLin}'s PyTorch integration is provided through \texttt{QuantumLayer}, a \texttt{torch.nn.Module} that exposes trainable circuit parameters $\mathbf{\theta}$, supports batching, and can be composed with other PyTorch modules. \texttt{QuantumLayer} is structured around three core concepts that govern how quantum information is represented and extracted. First, the \emph{measurement strategy} determines the layer output, such as full probability distribution (as obtained from hardware measurements), per-mode photon number expectations, or complex state amplitudes prior to measurement. Second, the \emph{computation space} specifies the Hilbert subspace used to represent the photonic state, ranging from the full Fock space to restricted or encoded spaces, including dual-rail or QLOQ qubit encodings~\cite{lysaght2024quantumcircuitcompressionusing}. The measurement model can further be specified through a detector model, which defines how photonic states are converted into classical outcomes, for instance using photon-number-resolving (PNR) or threshold detectors. Threshold detectors register the presence or absence of photons in an optical mode and are commonly used in experimental photonic QPU setups. The last concept is \emph{data encoding}, described in Section \ref{sec:data-encoding}. These concepts are illustrated in \texttt{Code Block} \ref{code:merlin_layer} and jointly decouple circuit evolution, state representation, and measurement semantics within \texttt{QuantumLayer}. 

\subsection{Data encoding}
\label{sec:data-encoding}

The encoding of classical data into quantum states remains one of the main bottlenecks in QML, with the potential to negate any potential quantum speedups if not addressed carefully \cite{schuld_effect_2021, gan_fock_2022}. 
\texttt{MerLin} supports two encoding strategies that are adapted to linear-optical hardware: \emph{angle encoding}, which maps classical features to phase shifts within the circuit, and \emph{amplitude encoding}, which directly initializes the quantum state amplitudes to match the input features. 
\subsubsection{Angle encoding}

Schuld et al. \cite{schuld_effect_2021} demonstrated that VQCs naturally realize a Fourier-like model: the input encoding fixes the accessible frequency spectrum (i.e., the feature basis), while the trainable circuit parameters only control how these fixed features are linearly combined. Gan et al. \cite{gan_fock_2022} generalized this framework to linear quantum photonic circuits, revealing a distinctive property of photonic platforms. Consider a parameterized circuit with $m$ modes, where data $x$ is encoded via a single phase shifter $S(x) = e^{i x \hat{n}_1}$ acting on the first mode (Figure \ref{fig:ql_circuit} with one phase shifter). The $n$-photon quantum model takes the form:
\begin{equation}
    f^{(n)}(x, \boldsymbol{\theta}, \boldsymbol{\lambda}) 
    = \sum_{\omega \in \Omega_n} c_\omega(\boldsymbol{\theta}, \boldsymbol{\lambda}) \, e^{i\omega x},
    \label{eq:photonic-fourier}
\end{equation}
where $\boldsymbol{\theta}$ parameterizes the trainable interferometers and $\boldsymbol{\lambda}$ the measurement observable. Crucially, the size of the frequency spectrum, and therefore the expressivity, increases linearly with the number of input photons $n$, without requiring additional encoding gates or deeper circuits. 
The \texttt{CircuitBuilder} API provides the \texttt{add\_angle\_encoding} method, which inserts  phase shifters at specified modes. Their phases correspond to classical features $\{x_i\}$, as demonstrated in \texttt{Code Block} \ref{code:merlin_layer} and Figure \ref{fig:ql_circuit}. A typical circuit choice consists of parametrized interferometers $W^{(i)}$ and an encoding layer $S(x)$ in between: 
\begin{equation}
    U(x, \boldsymbol{\theta}) = W^{(2)}(\boldsymbol{\theta}_2) \, S(x) \, W^{(1)}(\boldsymbol{\theta}_1).
\end{equation}

With \texttt{MerLin}, we successfully reproduced the results from~\cite{gan_fock_2022} using this type of circuits: Figure ~\ref{fig:ql_fourier} shows the optimal fit of a Fourier series using 3 modes and up to 3 photons, while Figure ~\ref{fig:ql_moon} illustrates the performance of the VQC on binary classification tasks.

\subsubsection{Amplitude encoding}

Amplitude encoding maps a classical data vector $\mathbf{x} \in \mathbb{C}^N$ directly into the amplitudes of a quantum state:
$\ket{\mathbf{x}} = \sum_{i=0}^{N-1} x_i \ket{i}$ where $\|\mathbf{x}\|^2 = 1$, and $\{\ket{i}\}$ denotes the Fock basis.

In \texttt{MerLin}, amplitude encoding is used when the input of the algorithm is a complex tensor or a \texttt{StateVector}. This encoding strategy is suited to scenarios where the quantum state is produced by an upstream process (such as another photonic circuit) rather than derived from raw classical features.


\subsection{Hardware-aware design}
\label{sec:hardware-aware}

Although \texttt{MerLin} is simulator-first, it is not hardware-agnostic. Its design reflects the constraints, capabilities, and evolution of photonic quantum hardware, enabling algorithm--hardware co-design. At the circuit level, \texttt{MerLin} operates directly on the native photonic components (Section \ref{sec:lo_background}), and users can select detector models and observables that match realistic readout strategies (Section \ref{sec:quantum-layer}). This hardware awareness extends to execution through the \texttt{MerlinProcessor} abstraction, which connects quantum layers to cloud-accessible photonic processors and allows parts of a hybrid model to be offloaded to real hardware, as illustrated in \texttt{Code Block}~\ref{lst:belenos}. This connector abstracts hardware-specific constraints such as latency, shot-based sampling, and limited parallelism, while preserving compatibility with modern machine learning workflows.

Beyond current accessible hardware, \texttt{MerLin} is designed to support the exploration of emerging quantum computing paradigms through its modular architecture. As an illustrative example, \texttt{MerLin} supports photonic quantum memristors: a memristor is a circuit element whose response depends on the history of current flow, mimicking synaptic plasticity. Its photonic analogue can be realized using a Mach-Zehnder interferometer, whose internal phases are updated based on detection statistics of a feedback loop. This mechanism introduces memory and nonlinearity, which are useful for tasks such as time-series prediction~\cite{selimovic_experimental_2025}.

Finally, \texttt{MerLin} acts as a bridge between computational modalities. Whereas qubit-based platforms operate in a $2^n$-dimensional computational basis, photonic processors naturally evolve in Fock space (Section \ref{sec:lo_background}). The \texttt{QuantumBridge} abstraction (see \texttt{Code Block} \ref{lst:quantumbridge}) provides a differentiable interface between these representations, enabling hybrid workflows across both paradigms. Beyond standard dual-rail encodings, \texttt{MerLin} supports general qubit-qudit mappings following the QLOQ scheme~\cite{lysaght2024quantumcircuitcompressionusing},
reducing photonic resource requirements and enabling systematic
cross-modality comparisons.

\begin{listing}[t]
\centering
\begin{minipage}{\linewidth}
\begin{codecellT}[QuantumBridge Usage] 
# Bridge: 2 qubits -> 2 photons, 4 modes (dual-rail)
bridge = ML.QuantumBridge(
    n_photons=2, n_modes=4,
    qubit_groups=[1, 1],
    computation_space=ML.ComputationSpace.UNBUNCHED
)
 
# Compose with upstream qubit circuit and downstream photonic layer
model = torch.nn.Sequential(qubit_circuit, bridge, merlin_layer)
\end{codecellT}
\end{minipage}
\caption{Quantum Bridge between gate-based circuit and photonic circuit, enabling interoperability between gate-based and photonic frameworks.}
\label{lst:quantumbridge}
\end{listing}

Overall, this hardware-aware design positions \texttt{MerLin} as a co-design tool rather than a purely abstract simulator: insights obtained from scalable (within simulable regimes), differentiable simulation inform both algorithmic development and the design of future photonic quantum processors.

\subsection{Complexity and simulability}

As mentioned earlier, simulating linear-optical quantum circuits is computationally challenging. While this observation underpins hardness arguments in photonic quantum computing, a growing body of work has shown that simulability can extend to larger-scale systems when additional structure or noise is present, as exemplified by results in Gaussian boson sampling and noisy linear-optical models. Such regimes, while classically tractable, remain highly relevant from a modelling and algorithmic perspective. \texttt{MerLin} deliberately embraces this viewpoint by focusing on strong simulation in controlled regimes where the full quantum state can still be computed exactly, or efficiently approximated. This enables access to complete probability distributions, exact expectation values, and analytic gradients via automatic differentiation capabilities.

From a methodological standpoint, the primary objective is discovery. This involves identifying which architectures, encodings, and training strategies lead to meaningful learning behavior. Considerations of simulability versus hardness are therefore treated as tools for interpreting and contextualizing these findings across classical and quantum regimes. In this sense, pushing simulation as far as possible is a deliberate choice that supports systematic investigation and informs subsequent studies in less tractable, hardware-based settings.

\section{MerLin as a benchmark-driven reproduction platform}
In this section, we summarize the reproduced papers, categorized either by task or method; a concise summary of results for each paper is provided in Table~\ref{tab:reproduced_papers} including confirmations, refinements, and observations on previously reported claims. 
To ensure proper benchmarking, models are ran several times and statistically sound results are provided, so that any user executing our code should observe results within the reported variance.
\begin{table*}[h]
\footnotesize
\centering
\begin{tabular}{|>{\RaggedRight\arraybackslash}p{0.270\textwidth}!{\vrule width 1.2pt}p{0.670\textwidth}|}
\hline
\textbf{Paper} & \textbf{Reproduction} \\
\noalign{\hrule height 1.2pt}
Fock State-enhanced expressivity of Quantum Machine Learning Models \cite{gan_fock_2022}
& We confirmed that, under the proposed encoding scheme, increasing the number of photons increases the expressivity of the variational quantum circuit. \\
\hline
Experimental quantum-enhanced kernel-based machine learning on a photonic processor \cite{yin_experimental_2025}
& We confirmed that accuracy improves with both training-set size and geometric difference. \\
\hline
Nearest Centroid Classification on a Trapped Ion Quantum Computer \cite{johri_nearest_2020}
& We reproduced this photon-native algorithm and obtained accuracies consistent with both the source paper and the classical baseline on the three considered datasets. \\
\hline
Experimental data re-uploading with provable enhanced learning capabilities \cite{mauser_experimental_2025}
& We confirmed that the fully quantum data reuploading model is both accurate and resource-efficient for binary classification on the four evaluated datasets, and that its expressivity increases with the number of reuploading layers. \\
\hline
Quantum Long Short-Term Memory \cite{chen_quantum_2020}
& Our MerLin-based photonic QLSTM matched the original gate-based QLSTM on function-fitting tasks, although the classical LSTM weaknesses reported in the source paper were less pronounced in our experiments. \\
\hline
Quantum Recurrent Neural Networks for Sequential Learning\cite{li_quantum_2023}
& We validated the partial measurement scheme in \texttt{MerLin} for temporal data prediction. \\
\hline
Photonic Quantum Convolutional Neural Networks with Adaptive State Injection \cite{monbroussou2025photonicquantumconvolutionalneural}
& By optimizing hyperparameters, we improved the reported binary classification accuracies from $92.7 \pm 2.1\%$ to $98.2 \pm 2.2\%$ on Custom BAS and from $93.1 \pm 3.6\%$ to $98.8 \pm 1.0\%$ on MNIST (0 vs 1). \\
\hline
Quantum Convolutional Neural Networks for classical data
classification \cite{hur_quantum_2022}
& We reproduced the reported advantage of QCNNs over CNNs under similar training budgets on a photonic architecture, but found that classical models remain more parameter-efficient. However, in photonic implementations this metric is of limited relevance, since the number of trainable parameters is inherently bounded by the number of phase shifters in the interferometer, while classical models saturate at lower absolute accuracy. \\
\hline
Quantum optical reservoir computing powered by boson sampling \cite{sakurai_quantum_2025}
& We reproduced the reported scalability, with performance improving as the number of modes increases, and confirmed that the model can be executed on QPU hardware. \\
\hline
Quantum Relational Knowledge Distillation \cite{liu_quantum_2025}
& We confirmed that quantum relational knowledge distillation yields a larger improvement in student performance than its classical counterpart. \\
\hline
Distributed Quantum Neural Networks on Distributed Photonic Quantum Computing \cite{chen_distributed_2025}
& We reproduced the parameter-efficiency gains obtained by using Boson Sampling to approximate CNN parameters, with training accuracy within approximately $2\%$, test accuracy within approximately $4\%$, and a fourfold reduction in required epochs, while also observing faster training with ADAM than with COBYLA; however, our ablation results differed from those in the source paper. \\
\hline
Quantum Self-Supervised Learning \cite{jaderberg_quantum_2021}
& The \texttt{MerLin} implementation outperformed the Qiskit version in both speed and accuracy, reaching $49.2\%$ accuracy on the first five CIFAR-10 classes after two epochs with eight modes, compared with $48.4\%$ for Qiskit and $48.1\%$ for the classical baseline. \\
\hline
Transfer learning in hybrid quantum-classical neural networks \cite{mari_transfer_2020}
& We obtained comparable accuracy across transfer settings and observed limited sensitivity to photon count. \\
\hline
Photonic quantum generative adversarial networks for classical data \cite{sedrakyan_photonic_2024}
& The \texttt{MerLin} implementation reproduced the original Perceval SSIM results while reducing training time by up to a factor of 15. \\
\hline
Computational Advantage in Hybrid Quantum Neural Networks: Myth or Reality? \cite{kashif_computational_2024}
& We confirmed that the HQNN achieves at least $90\%$ accuracy on the noisy spiral dataset with fewer parameters than a classical neural network across feature dimensions ranging from 5 to 60. \\
\hline
Quantum Large-Language Model Fine-Tuning \cite{kim_quantum_2025}
& Whereas the original paper reports up to a $3.14\%$ accuracy gain for the quantum-enhanced model, our reproduction found that the best quantum and classical models all reached approximately $89\%$ accuracy, with no clear separation between them. \\
\hline
Quantum Adversarial Machine Learning \cite{lu_quantum_2020}
& We confirmed that quantum classifiers remain vulnerable to both direct and transferred adversarial attacks, while adversarial training substantially improves robustness; on MNIST, we obtained approximately $98\%$ clean accuracy, $15\%$ BIM adversarial accuracy at $\epsilon=0.1$, and $95\%$ adversarial accuracy after adversarial training. \\
\hline
Experimental neuromorphic computing based on
quantum memristor \cite{selimovic_experimental_2025}
& We reproduced the main results and confirmed that adding a quantum memristor enhances the nonlinearity of the quantum reservoir, thereby improving performance on time-series tasks. \\
\hline
\end{tabular}
\caption{Reproduced papers used to validate MerLin's performance and utility. The main result of each reproduction is summarized here. Code and detailed results are available in the \href{https://github.com/merlinquantum/reproduced_papers}{MerLin/reproduced\_papers} repository.}
\label{tab:reproduced_papers}
\end{table*}

\subsection{Kernel methods}
Kernel methods compare data points through a similarity function
$k(x_1,x_2)=\langle\phi(x_1),\phi(x_2)\rangle$, implicitly embedding data into a feature space where linear models can capture non-linear structure. Quantum kernels extend this idea by encoding classical inputs into quantum states. In linear optics, data-dependent  unitaries $U(x)$ act on Fock states to produce embeddings $\ket{\phi(x)} = U(x)\ket{s}$, and a fidelity-based kernel is then given by $k(x_1,x_2)=|\langle \phi(x_1)\mid\phi(x_2)\rangle|^2$.

\texttt{MerLin} provides a scikit-learn--compatible implementation of fidelity kernels depicted in \texttt{Code Block} \ref{lst:kernel}. It also reproduces distance-based quantum kernels based on inner-product estimation \cite{johri_nearest_2020}. While originally demonstrated on qubit platforms using gates that emulate beam-splitter dynamics, these operations are native to photonic hardware. This reproduction validates \texttt{MerLin} and highlights its role as a bridge between abstract quantum kernel constructions and realistic photonic implementations.

\begin{listing}[t]
\centering
\begin{minipage}{\linewidth}
\begin{codecellT}[Fidelity kernel construction]
kernel = ML.FidelityKernel.simple(
    input_size=4, n_modes=6, n_photons=4)

K_train = kernel(X_train)
K_test  = kernel(X_test, X_train)
clf = SVC(kernel="precomputed").fit(K_train, y_train)
\end{codecellT}
\end{minipage}
\caption{Application of the FidelityKernel to a given training set.}
\label{lst:kernel}
\end{listing}
\subsection{Recurrent Architectures for Sequential Data}
Time series prediction is a fundamental paradigm in ML which involves learning temporal dependencies from sequential data to predict future values, traditionally addressed by recurrent architectures. While recent studies reveal that current QML models often struggle to outperform simple classical counterparts of comparable complexity for time series prediction \cite{fellner_quantum_2025}, they also highlight the potential of hybrid quantum-classical architectures when carefully designed. In particular, two works have applied variational quantum architectures to sequential learning. \cite{chen_quantum_2020} introduced the QLSTM, a hybrid model that replaces classical linear transformations within LSTM cells with variational quantum circuits. Subsequently, \cite{li_quantum_2023} proposed a hardware-efficient Quantum Recurrent Neural Network (QRNN) architecture featuring quantum recurrent blocks that preserve quantum information across time steps. We successfully implemented both models using \texttt{MerLin}, validating its use for temporal processing tasks, and providing tools for novel algorithmic development. 

\subsection{Convolutional Neural Network}
Convolutional Neural Networks (CNN) are foundational models for computer vision, owing to their locality, translation equivariance, and parameter sharing across feature maps \cite{he2015resnet}. Recently, \cite{Monbroussou_2025gatebasedqcnn} proposed a Subspace Preserving Quantum Convolutional Neural Network (QCNN) with a reported polynomial speedup over deep classical CNNs. This architecture was later adapted to photonic hardware \cite{monbroussou2025photonicquantumconvolutionalneural} using adaptive state injection, which implements pooling while preserving a fixed photon number and reducing the effective Fock-space dimension. We reproduced this photonic QCNN within \texttt{MerLin} using a density-matrix backend and, through a systematic hyperparameter study, improved the reported test accuracy on two datasets by $4–5\%$, thereby establishing a stronger baseline for future work. Beyond reproduction, \texttt{MerLin} natively supports amplitude-encoding with partial measurements. Combined with adaptive state injection, this enables a modular and transparent construction of photonic QCNNs using \texttt{QuantumLayer}, highlighting \texttt{MerLin}'s suitability for structured photonic QML models.

\subsection{Reservoir Computing}

The reservoir computing principle, introduced for processing temporal data \cite{jaeger2001echostatenetwork, maass2002liquistatemachine}, harnesses the complex dynamics of physical systems to map input data into a high-dimensional feature space. A linear probe takes this rich embedding as input. In the context of computer vision, popular models rely on a trained backbone, whether supervised \cite{he2015resnet} or self-supervised \cite{oquab2024dinov2}. By contrast, the reservoir is used as a fixed, untrained black-box system. Reservoir computing was adapted to linear optics in \cite{sakurai_quantum_2025, rambach_photonic_2025}. Ref. \cite{sakurai_quantum_2025} considers MNIST-like datasets. The circuit chosen is similar to the one in Figure \ref{fig:ql_circuit}. We successfully reproduced the key finding of this work: the accuracy scales with the size of the Fock space. Additionally, using \texttt{MerLin}, these experiments can be conducted both in simulation and on physical hardware through the \texttt{MerlinProcessor}, see Code Block \ref{lst:belenos}.

\subsection{Training paradigms for QNN}
Recent advances in QML have introduced several promising training paradigms that leverage quantum resources to enhance classical learning pipelines.  Quantum Relational Knowledge Distillation (QRKD) \cite{liu_quantum_2025} aligns teacher-student representations through quantum kernel functions, enabling geometric regularization in an exponentially large feature space while retaining classical inference.  Quantum Self-Supervised Learning (qSSL) \cite{jaderberg_quantum_2021}, originally implemented using \texttt{Qiskit}, combines contrastive learning with quantum neural networks to capture complex visual correlations in high-dimensional quantum feature spaces. The Distributed Quantum Neural Network (DQNN) framework \cite{chen_distributed_2025} uses photonic circuits as hypernetworks to generate classical network weights.

Using \texttt{MerLin}, we successfully reproduced the key results from these works. Our qSSL implementation exhibited improved scaling relative to the original Qiskit version, the DQNN reproduction achieved comparable MNIST accuracy, and QRKD consistently reduced the teacher--student performance gap relative to classical relational knowledge distillation across vision and language tasks.

\begin{listing}[ht]
\centering
\begin{minipage}{\linewidth}
\begin{codecellT}[Reservoir computation on Belenos QPU]
# 1) Create the Perceval RemoteProcessor (token needed)
rp = pcvl.RemoteProcessor("qpu:belenos")
# 2) Wrap it with MerlinProcessor
proc = ML.MerlinProcessor(
    rp,
    microbatch_size=32, # batch chunk size/cloud call
    timeout=3600.0)     # default wall-time/forward
# 3) Set your QuantumLayer to eval mode
reservoir.eval()
# 4) Run remotely
y = proc.forward(layer, X, nsample=5000)   # synchronous
\end{codecellT}
\end{minipage}
\caption{Reservoir computation running on remote Belenos QPU using \texttt{MerLin}.}
\label{lst:belenos}
\end{listing}

\subsection{Image generation}
Quantum circuits have been investigated as models for generative learning, notably for image generation. Quantum generative adversarial networks (QGANs) were among the first generative variational quantum algorithms to be designed \cite{Dallaire_Killoran_2018}, and a photonic QGAN was later proposed \cite{sedrakyan_photonic_2024}. The model is trained on the MNIST dataset in reduced dimension, using a patch-based approach, where each generator produces a patch of the total image. The generators are linear optical circuits whose output distribution is mapped to pixels, and the discriminator is a classical neural network. Using \texttt{MerLin}, we adapted the original \texttt{Perceval} code and simplified the training pipeline, switching from an SPSA-based optimization \cite{Spall_1998} to a faster PyTorch stochastic gradient descent optimizer.

\subsection{Benchmark Tasks for Model Expressivity and Robustness}
We evaluate \texttt{MerLin} on three challenging tasks. First, we replicate the quantum LLM fine-tuning results from \cite{kim_quantum_2025} on SST2 sentiment analysis; however, the binary classification setting appears too simple to draw definitive conclusions about quantum utility, a limitation consistent with concerns raised by Bowles et al. \cite{bowles_better_2024}. Second, following \cite{kashif_computational_2024}, we confirm that VQCs require fewer parameters than classical networks to achieve $\geq 90\%$ accuracy under increasing feature complexity. Third, investigating adversarial robustness \cite{lu_quantum_2020}, we find that amplitude encoding is highly vulnerable to small perturbations $\epsilon$, while angle encoding proves substantially more robust. Combining amplitude encoding with dimensionality reduction offers moderate protection but sacrifices the representational benefits of full amplitude encoding, highlighting encoding strategy as a critical design choice for adversarial settings.

\section{Discussion \& Conclusion}
In this work, we introduced \texttt{MerLin}, a quantum software platform designed for large-scale, simulation-based exploration of hybrid quantum--classical models while remaining explicitly hardware-aware. \texttt{MerLin} enables systematic investigation of quantum learning architectures within familiar PyTorch-based environments.

We validated the framework by reproducing a range of results from the QML literature. In doing so, we observed that implementations ported from \texttt{Perceval} to \texttt{MerLin} consistently benefited from
substantial improvements in simulation speed, reaching up to several order-of-magnitude speedups in some cases.
More importantly, for algorithms originally formulated in a gate-based setting and reformulated in a photon-native representation, we found that learning performance and qualitative behavior remained very similar. This indicates that photon-native models may effectively substitute gate-based models in many hybrid architectures, opening up a practical path both for translating existing gate-based QML literature to photonic platforms and for exploring cross-modality models within a unified framework.

Beyond individual reproductions, \texttt{MerLin} is designed to serve as a discovery engine for QML by promoting a benchmark-driven, empirical methodology inspired by modern machine learning practice. Through its open design and simplified reproduction framework, \texttt{MerLin} encourages the community to systematically benchmark new and existing results, study learning dynamics, and lower the entry barrier for both classical and QML practitioners. 

Finally, within the context of photonic quantum computing, \texttt{MerLin}'s hardware-aware design and support for hybrid execution position it as a co-design tool: insights obtained from large-scale simulation and benchmarking
can directly inform both algorithmic development and the design choices of
future photonic quantum processors. We believe this tight feedback loop between
software, algorithms, and hardware will be essential for identifying meaningful
near-term advantages in quantum machine learning.

\vspace{0.5em}
\textbf{Acknowledgements}\\
The authors thank Pierre-Emmanuel Emeriau for helpful discussions and insightful feedback during the development of \texttt{MerLin}, Hugo Izadi and Ankit Sharma for their contribution to the reproduced papers. 

\bibliographystyle{IEEEtran}
\bibliography{reference}

\end{document}